\pdfoutput=1

\documentclass[11pt, table]{article}

\usepackage{authblk}
\usepackage[final]{acl}

\usepackage{times}
\usepackage{latexsym}

\usepackage[T1]{fontenc}

\usepackage[utf8]{inputenc}

\usepackage{microtype}

\usepackage{inconsolata}

\usepackage{xspace}
\usepackage{color}
\usepackage{booktabs}
\usepackage{graphicx}
\usepackage{tcolorbox}
\usepackage{caption}

\usepackage{soul}
\usepackage{paralist}
\usepackage{enumitem}
\definecolor{lightblue}{RGB}{200,245,255}
\definecolor{lightred}{RGB}{255,200,200}
\definecolor{lightgreen}{RGB}{205,255,200}
\definecolor{lightgrey}{RGB}{225,225,225}
\definecolor{bggreen}{RGB}{204, 255, 153} 
\definecolor{bgblue}{RGB}{153, 204, 255} 
\definecolor{hlyellow}{RGB}{255,255,0}
\newcommand{\hlc}[2][yellow]{{\sethlcolor{#1}\hl{#2}}}

\newcommand{\nrf}{{NoReC\textsubscript{\textit{fine}}}\xspace}
\newcommand{\fen}{F\textsubscript{1}\xspace}
\newcommand{\ie}{\textit{i.\,e.}\xspace}
\newcommand{\eg}{\textit{e.\,g.}\xspace}

\newcommand{\wrt}{with respect to\xspace}

\title{Entity-Level Sentiment: More than the Sum of Its Parts
}

\author[1,2]{Egil Rønningstad}
\author[2]{Roman Klinger}
\author[1]{Lilja {\O}vrelid}
\author[1]{Erik Velldal}

\affil[1]{Department of Informatics, University of Oslo, Norway
}
\affil[2]{Fundamentals of Natural Language Processing, University of Bamberg, Germany}
\affil[ ]{\texttt{\{egilron, liljao, erikve\}@uio.no, roman.klinger@uni-bamberg.de
}}

\begin{document}
\maketitle
\begin{abstract}

In sentiment analysis of longer texts, there may be a variety of topics discussed, of entities mentioned, and of sentiments expressed regarding each entity. We find a lack of studies exploring how such texts express their sentiment towards each entity of interest, and how these sentiments can be modelled. 
 In order to better understand how sentiment regarding persons and organizations (each entity in our scope) is expressed in longer texts, we have collected a dataset of expert annotations where the overall sentiment regarding each entity is identified, together with the sentence-level sentiment for these entities separately.
  We show that the reader's perceived sentiment regarding an entity often differs from an arithmetic aggregation of sentiments at the sentence level.  %
Only 70\% of the positive and 55\% of the negative entities receive a correct overall sentiment label when we aggregate the (human-annotated) sentiment labels for the sentences where the entity is mentioned. 
Our dataset reveals the complexity of  entity-specific sentiment in longer texts, and allows for more precise modelling and evaluation of such sentiment expressions.

\end{abstract}

\section{Introduction}

\begin{table}[ht]
\centering
\small
\begin{tabular}{p{2cm}p{1.8cm}p{2cm}} %
\toprule
\rowcolor{lightgrey} \multicolumn{2}{l}{\textbf{Document level annotations}} &  \\
\textbf{Entity} &  & \textbf{Sentiment} \\
\rowcolor{bgblue}Mick Jagger & &Pos-Standard \\ 
\rowcolor{bggreen}Rolling Stones & & Pos-Slight \\
\midrule
\rowcolor{lightgrey} \multicolumn{3}{l}{\textbf{Sentence-level annotations}}\\
\textbf{Entity ref} & \textbf{Relation} & \textbf{Sentiment} \\
\rowcolor{bggreen} Rolling Stones& Mention& Neg-Slight\\
\rowcolor{bgblue} Mick Jagger & Mention & Neg-Slight \\
\multicolumn{3}{l}{(1) There is nothing pretty when \hlc[hlyellow]{Jagger} and} \\
\multicolumn{3}{l}{ the \hlc[hlyellow]{Rolling Stones} are on stage.} \\
\midrule
\textbf{Entity ref} & \textbf{Relation} & \textbf{Sentiment} \\
\rowcolor{bgblue} Mick Jagger & Mention & Pos-Slight \\
\multicolumn{3}{l}{(2) But \hlc[hlyellow]{Mick Jagger} knows what he is doing.} \\
\midrule
\textbf{Entity ref} & \textbf{Relation} & \textbf{Sentiment} \\
\rowcolor{bggreen} Rolling Stones& Coreference& Pos-Standard\\
\rowcolor{bgblue} Mick Jagger & member\_of & Pos-Standard \\
    \multicolumn{3}{l}{(3) Soon \hlc[hlyellow]{the band} delivers their unique rock'n roll} \\
        \multicolumn{3}{l}{aestethics that we came for.} \\
    \bottomrule

\end{tabular}
\captionof{figure}{Constructed example containing two entities and three sentences. The  document-level sentiment classifications on top are annotated separately from the sentence-level annotations. Sentence (1) contains mentions of both "Jagger" and "Rolling Stones". The mention of "Jagger" is resolved to "Mick Jagger", the most complete mention of that entity. Sentence (2) mentions "Mick Jagger" positively. 
Sentence (3) contains a sentiment regarding "the band". This is a coreference to "Rolling Stones". The annotators also classified the sentiment regarding "the band" to carry over to the entity "Mick Jagger" as member of that band.
}
\label{tab:toy2}
\end{table}

As the field of sentiment analysis progresses, sentiment analysis has developed from providing a single positive / negative polarity label for entire texts \citep[\eg][]{pang-lee-2004-sentimental}, into various fine-grained approaches, such as \textit{structured sentiment analysis}, where for each identified sentiment expression in a sentence, the sentiment category is classified, and the holder and target of the sentiment, if any, is identified \citep{barnes-2023-sentiment}. 
Often however, the end goal of sentiment analysis will be to extract more compound information about the sentiment expressed towards each entity.
Such overall sentiment classification per entity can facilitate for better media bias analyses and trend research where the source texts are more complex \citep{steinberger-etal-2017-large}. As we show in Section \ref{sec:rel_work}, we find few attempts to classify the overall entity-specific sentiment in longer texts.

To mitigate the lack of such entity-related sentiment data, we provide a Norwegian dataset of professionally written review texts annotated for sentiment both at the document and sentence level regarding each person and organization mentioned, \ie each \textit{volitional entity} in the text \citep{mitchell-etal-2013-open}. %
To our knowledge, our dataset is the first openly available of its kind, in any language, providing such separate sentiment labels for each entity, both at the sentence level and for the full document. Figure \ref{tab:toy2} exemplifies this multi-layered annotation scheme. It presents the annotation granularity at both document- and sentence level.

Our main contributions are as follows:
\begin{enumerate}[itemsep=1pt, topsep=2pt]  %
\item \textbf{A novel dataset and annotation scheme} for entity-wise sentiment classification both at the sentence- and at the full-text level, consisting of 412 texts containing 2479 entities.
\item \textbf{Analyses of the relations} between sentiments expressed locally (at the sentence level) and globally (at the full-text level) answering our research question (RQ1):  how consistently does sentiment towards each entity's mention agree with the entity's document-level sentiment?
\item \textbf{Classification of sentiment-relating sentences} We find that an important part of the sentiment signal regarding an entity is found in sentences where the entity itself is not the sentiment target. This answers our research question (RQ2): how can we quantify the gains from including a wider set of sentences than those containing a mention of the entity?
\item \textbf{Baseline models} for predicting the global sentiment based on sequence labeling and zero-shot LLM-prompting exemplify the complexity of the task. These are evaluated to a %
\fen of 56\% and 69\% respectively, and are described in Section \ref{sec:modelling}.
\end{enumerate}

\section{Related Work} \label{sec:rel_work}

We here present work and datasets that to various degrees support entity-specific sentiment classification for longer texts. Similar works on exclusively short texts are excluded, as these lack the complexity found in our dataset.

\textbf{Entities’ Sentiment Relevance Detection.} \citet{7395816} present and motivate the task of \textit{Entity-level Sentiment Analysis} (ELSA). We apply their task description of identifying the document-level sentiment per entity. Our works differs in that the main focus of their paper is identifying sentiment-relevant sentences for each entity.  They create a smaller dataset for the financial and medical domain. They do not describe the annotation process, and we find only 10 samples from each domain available on line today. We provide and describe a larger dataset, and the focus of our modelling is the end goal of identifying the entity-level sentiment at the document-level.

\textbf{Document-level Sentiment Inference.} \citet{choi-etal-2016-document} aim at inferring not only a sentiments expressed regarding each entity in the text, but also the holder of each sentiment conveyed in newsmedia texts. Their suggested model for this demanding task evaluates to well below 50\% \fen on all evaluations reported.  In our work, the holder is understood to be the author of the text, and we focus on the sentiment relations between different entities and references in the text via both coreferential and other anaphoric relations.

\textbf{PerSenT.}  \citet{bastan-etal-2020-authors} annotate documents for one entity each, the main person of interest in the text, both at the document- and the sentence-level.  \citet{Kuila2024DecipheringPE} employ this dataset in their task of determining the overall sentiment polarity expressed towards a target entity in news texts. The PerSenT dataset is annotated by crowd-sourcing, annotated for only one entity per text, and the text length is limited to 16 sentences. In contrast, our dataset annotates the texts for all volitional entities mentioned in the text. It is annotated and curated by trained individuals, and the texts contain on average 27.5 sentences. 

\textbf{NewsMTSC.} This dataset by \citet{hamborg-donnay-2021-newsmtsc} and the subsequent multilingual {MAD}-{TSC} \citep{dufraisse-etal-2023-mad} contain news texts with each sentence labeled for sentiment regarding important volitional entities mentioned by name in the sentence. The entities are given identifiers that allow for sentiment aggregation, but an overall sentiment per entity and text is not identified. {MAD}-{TSC} contains 4714 sentences regarding 1007 labeled entities, with an average sentence length of 31 words.

\textbf{ELSA-pilot}. In \citet{ronningstad-etal-2022-entity}, we presented a pilot study  that motivates treating the global sentiment separately from the local sentiments. Crucially, we found that aggregation of sentence-level sentiment scores do not sufficiently capture the entity-dependent signals regarding the overall sentiment. 
 We find that in the texts inspected, sentiment is related to entities not only through name mentions and coreferences, but through sentences with other relations as well. The findings were exploratory,  and not supported by a more complete dataset.

\section{Data Collection}\label{sec:annotation}

The Norwegian Review Corpus \citep[NoReC,][]{velldal-etal-2018-norec} %
contains 43,436 professional Norwegian newspaper reviews from a range of domains, such as music, literature, restaurants, movies, electronics and more. The reviews typically balance both positive and negative assessments of the entity under review as well as various background information. 

The \nrf \citep{ovrelid-etal-2020-fine} corpus contains a subset of 412 reviews from the NoReC corpus. These texts are annotated for fine-grained sentiment information, including holders, polar expressions, polarity, and intensity. %
We chose this dataset as our texts, and enrich the dataset with new entity-focused sentiment annotations. More details on our dataset can be found in Table \ref{tab:total_counts}.
\subsection{Pre-processing}  
Since our task is to annotate texts for sentiment towards individual volitional entities, we trained a dedicated named entity recognition (NER) model for Norwegian on the NorNE dataset \citep{jorgensen-etal-2020-norne}, but included only the PER and ORG labels (merging GPE-ORG with the ORG category). 

All mentions of an entity were clustered through substring matching, to obtain a list of entities per documents and their mentions. If an entity "John Travolta" was mentioned in a text, the mentions "John" and "Travolta" would be clustered together with "John Travolta". There is in Norwegian little case inflection of proper nouns, besides genitive where the characters "'" and/or "s" are added. Our substring matcher would therefore check if stripping of "s" and "'" would give match. This way, "John's" would be found to be a substring of "John". We found few clustering errors from this approach. One exception was that "Elisabeth I" was found to be a substring of "Elisabeth II", and the two we therefore clustered and counted as one entity in the text.

\begin{table} %
  \centering
  \small
  \newcommand{\nn}{\phantom{0}}
\begin{tabular}{@{}lcccc@{}}
\toprule
Split &  Entities &  Texts &  Sentences &  Annotations \\
\midrule 
Test      &   \nn247 &     \nn44 &     \nn1252 &       1057\\
Train  &  2232 &    368 &    10083 &       8834 \\
\midrule
Sum       &  2479 &    412 &    11335 &       9891 \\
\bottomrule
\end{tabular}
    \caption{Total counts of texts, entities, sentences and annotations for the dataset after cleaning and postprocessing, as per its initial release.}
    \label{tab:total_counts}
\end{table}

\subsection{The annotation task}

For each volitional entity in each document, the task of the annotators is to annotate sentiment at two different levels, as exemplified in Figure \ref{tab:toy2}:
\begin{enumerate}[leftmargin=0pt, nosep, itemsep=1pt, topsep=2pt]
    \item[] \textbf{Document level:} Based on a reading of the entire text with the given entity in mind, label  the sentiment that the full text conveys towards that entity.
    \item[] \textbf{Sentiment-relevant references:} For each sentiment-relevant sentence, identify the text span that either directly refers to the entity in question or indirectly contributes to the entity-directed sentiment through a specified semantic relation. The possible relations are, in order of priority:   
    
    \begin{enumerate} [ itemsep=1pt, topsep=2pt]
        \item Name mentions, \eg "Jagger", "Rolling Stones". 
        \item Coreferences, \eg "they", "the band".  
        \item Bridging references. In addition to coreference, we annotate anaphoric relations between entities that are not co-referent, so-called bridging relations. The inventory of relations was motivated by the pilot study described in \citet{ronningstad-etal-2022-entity} and included the relations "member\_of", "has\_member" and "created\_by". When other bridging relations implied sentiment regarding a target, this was annotated under the subsequent point.
        \item Whenever a sentence was considered to imply sentiment regarding an entity in any other way than the above mentioned, the entire sentence was labelled with sentiment, but no text span inside the sentence was identified. 
    \end{enumerate}
\end{enumerate}

The relation categories for the bridging relations were suggested to the annotators from our initial exploration of the data, and in annotation meetings it was established that these categories were relevant and sufficient for the dataset at hand. See the annotation guidelines' list of terms in Appendix \ref{sec:appendix_guidelines} for further description of coreferences and bridging references.

All sentiment annotations employ a five-category scale, similar to  \citet{dufraisse-etal-2023-mad} and \citet{bastan-etal-2020-authors}: "Negative--Standard", "Negative--Slight", "Neutral", "Positive--Slight", and "Positive--Standard". 
For the "Neutral" category, only name mentions are identified, since these could be added in the pre-processing. The other references to an entity were only annotated if they were non-neutral. Annotation was carried out using the Inception tool \citep{klie-etal-2018-inception}.  Figure \ref{fig:inception} in Appendix \ref{sec:appendix_annotation_example} shows example screenshots from the annotation process.

\subsection{Annotation guidelines}\label{sec:guidelines}
 Our annotation guidelines are derived from those of \nrf, which in turn build on the work of \citet{Kauter2015TheGT}. An English translation of the guidelines is presented in Appendix~\ref{sec:appendix_guidelines}, and  we briefly present some of the most central considerations below. 
\paragraph{When factual statements express sentiment.}
Our guidelines conclude that "pure" factual statements without any indication of sentiment from the author, should be considered neutral. One should limit the need for  domain knowledge from outside the discourse, in order to conclude whether a piece of information should be classified as conveying any sentiment polarity. According to these rules, the sentence "The Rolling Stones album sold over 22 million copies." contains no sentiment towards "The Rolling Stones". 

\paragraph{When sentiment towards related targets implies sentiment towards the volitional entity.}
If the annotator perceives a sentiment expression towards a movie to imply sentiment towards the director, the annotator would, when annotating \wrt the director, label the movie as "created by", and label the sentiment that this related target would have. Each case requires separate consideration by the annotators. %

\paragraph{Annotate the most prominent reference.} If an entity has more than one reference in a sentence, we annotate the name mention before coreferences, and coreferences before other anaphoric references. In the sentence "John played for us and we all love him.", the name mention "John" would be annotated with positive sentiment, although the sentiment expression "love" has "him" as target, a coreferent to "John".\\ %

\subsection{Annotation process} \label{sec:process}
 The dataset was annotated by five paid NLP students at the BSc level. All are native Norwegian speakers between 20 and 35 years old. They underwent introductory training and test annotations in preparation for the project. During this introductory training, the annotators contributed towards refining the annotation scheme. All annotations were curated by the first author of this paper, as the project leader.

After manual cleaning, the pre-annotated volitional entities, 2481 documents based on 412 texts remained for further analysis. Final counts for the dataset are presented in Table~\ref{tab:total_counts}. 
All annotators took part in the three phases of the project: 
\begin{compactenum}
  \item \textbf{Introductory parallel annotations and discussions.}
Annotators were initially provided with 75 documents, whereby 2--3 annotators would annotate the same texts. The annotators then inspected each others' work, the guidelines were discussed and if neccessary adjusted.
\item \textbf{The entire training corpus annotated.}
The annotators subsequently annotated the 2481 document in the dataset, according to availability, one annotator per document. Each annotator annotated from 200 documents and upwards. The number of documents annotated by each annotator is shown in Table~\ref{tab:sentlevel_3cats_cohen}. 
\item \textbf{Parallel annotation of the test set.}
Finally, all annotators annotated the test split, as pre-defined in \nrf, in parallel. The test data contains 44 different texts, containing a total of 247 volitional entities. Each text contains on average 28.5 sentences.
\item \textbf{Curation} The project leader reviewed all annotations in the dataset. For the training and development splits, there was one annotator to review. The annotations were corrected when neccessary. The amount of document-level annotations corrected by the curator, varied among the annotators from 0.5\% to 8.2\%. For the test split, all annotators annotated all instances. The curator inspected the majority vote before making the final judgement. The agreements here are shown in Tables \ref{tab:cohen_kappa} and \ref{tab:sentlevel_3cats_cohen}. 

\end{compactenum}

\subsection{Annotator agreement} \label{sec:agreement}
We present here the annotator agreements, both for the overall sentiment per entity, and for the sentence-level annotations.  For these analyses, we remove the intensity levels "Slight" and "Standard", and check for agreement only in terms of the main categories "Positive", "Neutral" and "Negative". %

\paragraph{Document--entity sentiment.}
We first inspect annotator  agreement for the overall sentiment assigned to each volitional entity at the document-level. Table~\ref{tab:cohen_kappa} shows the agreement towards the curated version. We find that the mean Cohen's Kappa among annotators compared to the curated document labels was 0.71, and standard deviation among the five annotators is 0.11. \\ %
\paragraph{Sentence--entity sentiment.}
We then turn to annotator agreement at the sentence-level, again with respect to the labels "Positive", "Neutral" or "Negative", with Cohen's kappa shown in  Table~\ref{tab:sentlevel_3cats_cohen}. Mean Cohen's kappa for agreement with the curated annotation is 0.72, and standard deviation among the annotators is 0.06.

\begin{table}[t]
    \centering
    \small
\begin{tabular}{@{}llllll@{}}
\toprule
{} &  ann\_1 &  ann\_2 &  ann\_3 &  ann\_4 &  ann\_5 \\
\midrule
curated        &  0.53 &  0.81 &   0.75 &  0.67 &  0.80 \\
ann\_1          &    1.0 &  0.43 &  0.34 &  0.35 &  0.41 \\
ann\_2          &        &    1.0 &  0.65 &   0.66 &  0.79 \\
ann\_3          &        &        &    1.0 &  0.60 &  0.71 \\
ann\_4          &        &        &        &    1.0 &  0.69 \\
\midrule
\# ann'd &    380 &    820 &    515 &    245 &    875 \\
\bottomrule
\end{tabular}
    \caption{Cohen's kappa agreement on the documents' sentiment polarity for each entity. Mean agreement with the curated result is 0.71.   "\# ann'd" indicate how many documents in the dataset each annotator had annotated before starting on the test set.}
    \label{tab:cohen_kappa}
\end{table}

\begin{table}[t]
    \centering
    \small
\begin{tabular}{@{}lllllll@{}} %
\toprule
{} &  ann1 &  ann2 &  ann3 &  ann4 &  ann5 \\
\midrule
curated &    0.64 &  0.77 &  0.78 &   0.68 &  0.74 \\
ann\_1   &        1.0 &   0.54 &  0.57 &   0.52 &  0.53 \\
ann\_2   &    &  1.0 &  0.71 &  0.65 &    0.70 \\
ann\_3   &         &        &    1.0 &  0.63 &  0.77 \\
ann\_4   &           &        &        &    1.0 &  0.65 \\
ann\_5   &           &        &        &        &    1.0 \\
\midrule
\# ann'd &  380 &  820 &  515 &  245 &  875 \\
\bottomrule
\end{tabular}

    \caption{Cohen's kappa agreement between annotators and the curated conclusion for sentiment polarity on the sentence level, \wrt the given entity. Mean annotator agreement with curated is 0.72.}
    \label{tab:sentlevel_3cats_cohen}
\end{table}

\paragraph{Conclusions from analyzing inter-annotator agreement.}
Despite individual variations in agreement, mean Cohen's kappa agreement at both the document- and sentence level is above 0.70. We consider this to be a satisfactory level of agreement, and an indication that the annotators indeed were able to identify and classify the requested sentiment signals in the texts.
Inspecting selected disagreements indicate that one source of disagreement lies in  drawing the line for how much world knowledge to include in a sentiment judgement. \citet{zaenen-etal-2005-local} argue that world knowledge underlies just about everything we say or write, and that this leads to diverging readings of a text. We found in our data that annotators in deed tended to disagree, \eg when a person commonly considered to have been "good" or "bad" was mentioned without a particular sentiment expressed in the text. During curation, these cases would be judged as Neutral.

We further find it noteworthy that the two annotators with the fewest documents annotated have the lowest agreement with the curated version. The minimum requirement was to annotate at least 200 documents before proceeding to annotating the test set. But in our case, annotators who annotated more than 400 documents had noticeable higher agreement with curated, as seen in Tables \ref{tab:cohen_kappa} and \ref{tab:sentlevel_3cats_cohen}.

\section{Dataset Analysis}\label{sec:anylsis-main}
We here present selected analyses of the main body of the dataset.  The results in this section relate to the training and development splits combined. This collection contains 368 reviews with 2232 volitional entities and 8834 sentiment labels in total. The focus is on providing answers to RQ1 about the relations between the individual mentions of an entity and the entity's document-level sentiment. We also answer RQ2 by quantifying the gain from including more sentences than those containing a mention of the entity.

\begin{table}[t]
\centering    
\small

\begin{tabular}{@{}lrr@{}}
\toprule
Relation category & {\#} & {\%} \\
\midrule
Name mention & 1382 & 36.7 \\
Coreference, anaphoric & 432 & 11.5 \\
Bridging: created\_by & 966 & 25.7 \\
Bridging: has\_member & 294 & 7.8 \\
Bridging: is\_member & 48 & 1.3 \\
Sentence-level sentiment & 641 & 17.0 \\
\midrule
Total & 3763 & 100.0 \\
\bottomrule
\end{tabular}

\caption{All non-neutral sentiment annotations in the training and development split of the dataset. We find that only 36.7\% of the annotated sentiments are on sentences containing an entity's name mention.  %
}
\label{tab:cats-sentiments}
\end{table}

\begin{table*}[t]
    \centering
        \small
\begin{tabular}{@{}lrrrrrrr@{}}
\toprule
mentions\_only &  Neg-Std &  Neg-Slight &  Neutral &  Pos-Slight &  Pos-Std &  Total &  Neutral  \\
Document-level &          &             &          &             &          &        &            pct  \\
\midrule
Positive--Standard                &        1 &           4 &      134 &          32 &      502 &    673 &         19.9 \\
Positive--Slight                  &        0 &           4 &       43 &          54 &       17 &    118 &         36.4 \\
Neutral                          &        2 &           2 &     1139 &          12 &        4 &   1159 &         98.3 \\
Negative--Slight                  &       10 &          59 &       46 &           1 &        5 &    121 &         38.0 \\
Negative--Standard                &       97 &           9 &       48 &           5 &        2 &    161 &         29.8 \\
\bottomrule
\end{tabular}
    \caption{Sentiment towards entities' name mention vs. sentiment towards the entity at the document level. Sentiments at the name mention level are aggregated by averaging the non-neutral sentiments. When inspecting the "Neutral" row, we find that 1139, or 98.3\% of the Neutral entities in the documents, had neutral sentiment towards all entity mentions. For the entities with sentiments, we find that 19.9--38 pct of these had no sentiment at the name mentions, and were incorrectly  aggregated to "Neutral".}
    \label{tab:mention_document}
\end{table*}

 \begin{table*}[bht]
    \centering
    \small
\begin{tabular}{@{}lrrrrr@{}} %
\toprule
{Aggregated entities} &  mentions &  mentions &  mentions &  all sentiments &  support \\
{} &   &   coreferences &  coreferences &   &   \\
{Document-level} &   &   &   bridging &   &   \\
\midrule
Positive--Standard &                        83.5 &                  84.9 &                       89.0 &           89.5 &    673\\
Positive--Slight   &                        48.6 &                  46.9 &                       48.3 &           48.8 &    118 \\
Neutral           &                        88.7 &                  90.1 &                       93.6 &           94.5 &   1159 \\
Negative--Slight   &                        59.3 &                  55.8 &                       64.4 &           63.2 &    121 \\
Negative--Standard &                        71.6 &                  73.4 &                       75.4 &           77.4 &    161 \\
Accuracy          &                        82.9 &                  83.7 &                       86.6 &           87.2 &       \\
Weighted avg      &                        82.2 &                  83.2 &                       86.9 &           87.6 &       \\
\bottomrule
\end{tabular}
    \caption{  \fen scores for the five sentiment classes in the dataset, when using increasingly more of the annotated data. For the first column, only sentiments directed towards the entity mentions are aggregated. For the next column,  coreferences are added. Then targets with a bridging relation are added, before all annotations at a sub-document level as aggregated per entity. All aggregations use the strategy of averaging non-neutral sentiments.
  This table is graphed in Figure~\ref{fig:doc-level_improvements}. }
    \label{tab:mention_anaphoric_prf}
\end{table*}

\begin{figure}[t]
    \centering
    \includegraphics[width=\columnwidth]{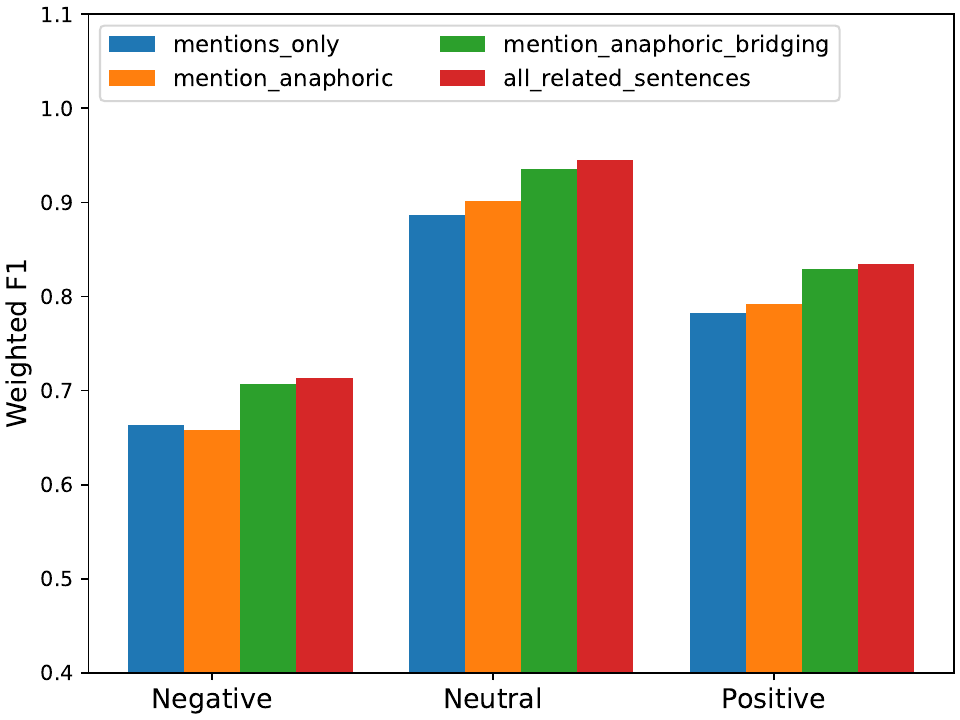}
    \caption{Improvements per sentiment category and sentence relation category. We here use a weighted average of the two sentiment intensities in Table~\ref{tab:mention_anaphoric_prf}. %
    }
    \label{fig:doc-level_improvements}
\end{figure}

\subsection{Annotations and polarity-counts}
Table~\ref{tab:cats-sentiments} shows the distribution of annotations in the dataset, across category and sentiment. It shows that the sentiment-relevant coreferences beyond name mentions are comparably few. In contrast, we find that the bridging relations (created\_by, has\_member and is\_member) contribute quite significantly to entity-directed sentiment in our dataset. These relations constitute 34.8\% of the annotated sentiments, almost equally frequent to the sentiment labels that are directly attached to an entity mention. These figures indicate that any approach that labels only sentences for sentiment regarding an entity if that entity is named in the sentence, appear to lose the majority of sentiment signal, which is found in sentences with other relations to the entity.

\subsection{Document-level vs aggregated lower level sentiment} \label{sec:doc_agg}
In this work we are specifically interested in the relations between the high-level and lower-level annotations for each entity. The availability of our dataset enables further analysis of these correlations at the per-entity level. In the following we will attempt to evaluate the effect of 
each category of sentiment-related sentences and how the aggregation of lower levels of sentiment classifications compare to the document-level score independently assigned by our annotators.
We start by aggregating the sentiments for the name mentions only, before we add the remaining available annotations. When referring to the aggregated sentiment score, we here refer to averaging and rounding the non-neutral mentions, whereby we assign the "Standard" sentiments the value of $\pm$2, and slight sentiments are $\pm$1. %

\paragraph{Sentiment towards entity mentions only.} 
In the previous section, we established that the majority of sentiment signals in our texts lay in sentences without the entity explicitly mentioned. However, if the sentiment signals from the sentences containing an entity mention are coherent with the sentiment signals in sentences with other relations to the entity, these latter sentences would be redundant in order to satisfactorily locate the document's overall sentiment regarding the entity in question. 
Table~\ref{tab:mention_document} shows the confusion matrix for aggregated sentiment for name mentions, compared with the annotated document-level sentiment. The "Neutral" column shows the distribution of entities that do not receive any sentiment towards their name mention, over their true, document-level sentiment. We see that 19.9\% of the true "Positive--Standard" entities receive no sentiment towards their name mention, while 38\% of the true "Negative--Slight" entities are without any sentiment towards their name mention. This gives an answer to RQ1 through the observation that 271 out of the 1073 non-neutral entities in the dataset (25\%) are incorrectly assigned a neutral sentiment by the sentiment-bearing sentences where their name is mentioned. To correctly classify these entities, we need to find a sentiment signal in other parts of the text. 

\paragraph{Sentiment towards name mentions and references.} 
In order to further understand the sentiment contributions of the various references to an entity, we compare the \fen scores for aggregated sentence-level gold sentiment labels. We start with the sentences with name mentions only, gradually adding more sentiment relation categories. This may be considered an ablation study where we explore the impact of the various parts of the dataset's categories. 
We start with the name mention sentiments, as described in the previous subsection. Subsequently, the coreferences are added, then the bridging mentions, and finally the sentence-level sentiment annotations. Table~\ref{tab:mention_anaphoric_prf} and Figure~\ref{fig:doc-level_improvements} shows that aggregating sentiment expressed towards both name mentions and anaphoric coreferences add just one percentage point to the support-weighted average \fen.  Adding the targets with a bridging relation to the entity, though, improves the average \fen by an additional 3.7 percentage points. %
From there, including also other sentences where the annotators found a sentiment-relevant relation to the entity, only improves weighted average \fen from 86.9\% to 87.6\%.\\ 
These findings indicate an answer to RQ2, that in order to find the sentiment-relevant parts of a text \wrt an entity, looking only at sentences with an entity's name mention or even including any anaphoric coreference to the entity, is not enough. Having a model that can also capture sentiment from sentences where a target has a bridging relation to the entity, appears to be important.

\begin{table}[t]
    \centering
    \small 
\begin{tabular}{@{}lrrr@{}}
\toprule
Entity  &                                &  non-  &   \\
mentions &              neutral &         neutral   &      Total           \\
\midrule
Multiple           &       231 &             566&   797\\
Single            &       928 &             507&  1435 \\
\bottomrule
\end{tabular}
    \caption{Distribution of neutral and non-neutral entities, with one or multiple name mentions in the text. 507 out of the 1435 entities mentioned only once, receive a non-neutral sentiment. }
    \label{tab:single_neutral}
\end{table}
\subsection{Are single mentions in general neutral?}
In our dataset, almost two thirds of the entities are mentioned only once by their name in a given text. If we, as suggested by \citet{dufraisse-etal-2023-mad} could assume that entities mentioned only once are neutral and not in focus, that would simplify the task considerably. For our dataset, Table~\ref{tab:single_neutral} shows that although a majority of the entities with only one name mention are neutral, nearly half of the entities receiving a polarity are single mention entities. Discarding these would have meant discarding much valuable sentiment information, and we conclude that entities with one name mention only are worth keeping. 

\section{Baseline Modelling}
\label{sec:modelling}
We here present two approaches to using language models for predicting the document-level sentiment regarding each entity mentioned in the text. Due to the richness of annotations, neither of these utilize all available annotations in the dataset. The first approach fine-tunes a model for finding the relevant entity mentions and labeling these with sentiment polarity "Positive", Negative" or "Neutral". The heuristics described in Section \ref{sec:doc_agg}
 aggregates these to the document-level prediction. The second approach prompts a large language model with the text, the entities, and a request to return the document-level sentiment label for each entity.
 
\subsection{Predicting and aggregating mentions' sentiments}
We extract a simplified dataset containing only the entity mentions and their sentence-level sentiments.  We train a sequence labeler to identify entities and their three-class sentiment, with evaluation results shown in Table~\ref{tab:elsa_tsa_mod}. The pretrained model applied was NorBERT3-large\footnote{\url{https://huggingface.co/ltg/norbert3-large}} \citep{samuel-etal-2023-norbench}. The models tested and search space for hyperparameters explored are shown in Table \ref{tab:seq_label_params} in Appendix~\ref{sec:appendix_baseline}.

As discussed in Section \ref{sec:anylsis-main}, the document-level sentiment can not be fully derived from the set of sentiments regarding each mention of an entity. However, we aggregate the predicted sentiments, similarly to how we aggregated the annotations for each entity mention in Section \ref{sec:anylsis-main}. This approach serves as a naive modelling baseline and an example of the limitations of this approach.
The results from aggregating the modelled labels to the document-entity level are presented in Table \ref{tab:elsa_tsa_agg}.  Table \ref{tab:mention_anaphoric_prf} shows that 82.9\% of the entities in the training and development splits were correctly classified at the document level when aggregating the true sentiment labels for the entities' mentions, and serve as an upper bound for this approach. Table \ref{tab:elsa_tsa_agg} shows that when aggregating the predicted labels, 70.9\% of the entities in the test split were correctly labeled with this baseline model.

\begin{table}[bht]
    \centering
    \small
\begin{tabular}{@{}lrrrr@{}}
\toprule
 & Precision & Recall & \fen & Support \\
\midrule
Neg & 70.6 & 41.4 & 52.2 & 29 \\
Neu & 73.9 & 88.3 & 80.5 & 308 \\
Pos & 68.0 & 57.1 & 62.1 & 119 \\
Macro avg & 70.8 & 62.3 & 64.9 & 456 \\
W. avg & 72.2 & 77.2 & 73.9 & 456 \\
\bottomrule
\end{tabular}
    \caption{Sequence labelling of each individual entity name in the test split. An exact match for both the text span and sentiment label is required for the predictions to be counted as correct. At this level there is no aggregation. Aggregated sentiment labels per entity are presented in Table \ref{tab:elsa_tsa_agg}.}
    \label{tab:elsa_tsa_mod}
\end{table}

 \begin{table}[bht]
    \centering
        \small
\begin{tabular}{@{}lrrrr@{}}
\toprule
 & Precision & Recall & \fen & Support \\
\midrule
Neg & 44.4 & 19.0 & 26.7 & 21 \\
Neu & 67.4 & 95.5 & 79.0 & 132 \\
Pos & 88.2 & 47.9 & 62.1 & 94 \\
Accuracy & 70.9 & 70.9 & 70.9 &  \\
Macro avg & 66.7 & 54.1 & 55.9 & 247 \\
W. avg & 73.4 & 70.9 & 68.1 & 247 \\
\bottomrule
\end{tabular}
    \caption{Aggregated sequence labels from the baseline sequence labeling model, evaluated against the entities in the test split.}
    \label{tab:elsa_tsa_agg}
\end{table}

\subsection{Zero-shot LLM prompts}
Recent work indicates that ChatGPT and open-source counterparts may be a relevant resource for annotating and labeling English texts \citep{Gilardi2023ChatGPTOC, Alizadeh2023OpenSourceLL}. We therefore constructed a zero-shot dialogue with ChatGPT.

The prompts were what we consider clear and well-posed Norwegian questions about which of the three sentiment categories "Positive", "Neutral" or "Negative" is assigned to a given entity by the text. We performed the dialogue through the web interface with a paid monthly subscription to OpenAI, employing GPT v4 \citep{OpenAI2023GPT4TR}. %

The initial prompt was the entire text, preceded with this sentence in Norwegian: 
"In the subsequent text, is the sentiment towards "Kirsten Flagstad" Positive, Negative or Neutral?"\\
Where "Kirsten Flagstad" is the volitional entity in question. The prompt would be a lengthy answer including reasoning.
The next prompt would be, translated: "Please give the answer with one word, Positive, Negative or Neutral". Table \ref{tab:doclevel_gpt} shows that this zero-shot approach yielded an accuracy of 73.3\%

\begin{table}[t]
    \centering
    \small
\begin{tabular}{@{}lrrrr@{}}
\toprule
 & Precision & Recall & \fen & Support \\
\midrule
Neg & 60.0 & 57.1 & 58.5 & 21 \\
Neu & 77.4 & 72.7 & 75.0 & 132 \\
Pos & 70.9 & 77.7 & 74.1 & 94 \\
accuracy & 73.3 & 73.3 & 73.3 &  \\
macro avg & 69.4 & 69.2 & 69.2 & 247 \\
W. avg & 73.4 & 73.3 & 73.3 & 247 \\
\bottomrule
\end{tabular}
    \caption{
    Predicted sentiment labels per entity at the document level in the test split, provided through ChatGPT with GPT-4.
    }
    \label{tab:doclevel_gpt}
\end{table}

\section{Conclusion} \label{sec:conclusion}
We have presented a dataset annotated for entity-level sentiment analysis based on professional review texts in Norwegian. The dataset allows for training and evaluating models for entity-wise sentiment analysis. We have shared insights from the dataset creation, and analyzed how sentence-level expressions of sentiment regarding an entity relate to the entity's overall document-level sentiment. 
The dataset is available online.\footnote{\url{https://github.com/ltgoslo/ELSA.git}}

\section*{Acknowledgements}
The work documented in this publication has been carried out within the NorwAI Centre for Research-based Innovation, funded by the Research Council of Norway (RCN), with grant number 309834.\\
Baseline model training was performed on resources provided by
Sigma2 -- the National Infrastructure for High-Performance Computing and
Data Storage in Norway.\\
We would like to thank the anonymous reviewers for their helpful comments. We are thankful for the skilful annotation work and overall contributions from the research assistants who annotated the dataset: Daniel Skinstad Drabitzius, Iunia Melania Antal, Birk Søråsen, Ellen Margrethe Ulving, and Håkon Liltved Hyrve.

\section{Ethical Considerations} \label{sec:ethics}
We are not aware of any misconduct or violation of rules and regulations during the work with the presented dataset. The newspaper reviews used for our dataset are previously published and made available for research. The annotators were compensated with the university's standard wages as research assistants for all hours of involvement in the project. 

The reviews in our dataset are sampled from a corpus of Norwegian reviews published in the periode 2003--2017.  Opinions and writing styles could be considered representative for their news sources and time period.

The raw texts in our dataset have been publicly available for several years, and have been available for llms to train on. Our annotations were not publicly available when we prompted ChatGPT, and were not submitted as examples during our ChatGPT experiments.

\bibliography{custom}

\appendix
\section{ Annotation Example}\label{sec:appendix_annotation_example}
Figure \ref{fig:inception} in the appendix shows segments from screenshots of one text being annotated for two different entities, two members of the same band: \textit{Julian Casablancas}, the band leader and vocalist, and \textit{Nick Valensi}, the guitarist in the band. The text is machine translated from Norwegian and just briefly corrected. Green labels are at the segment level: The sequence is annotated for relation to the entity and for sentiment. Blue labels are the overall, document-level sentiment towards an entity. Each document therefore, has one such annotation. Pink labels are for sentences expressing sentiment with an unspecified relation to the entity. We see that the annotators found some sentiments towards the band to imply sentiment towards the band leader Julian Casablancas. For the guitar player Nick Valensi, only sentiment regarding him directly was recorded. 

\begin{figure*}[hb!]
	\centering
\includegraphics[width=\textwidth]{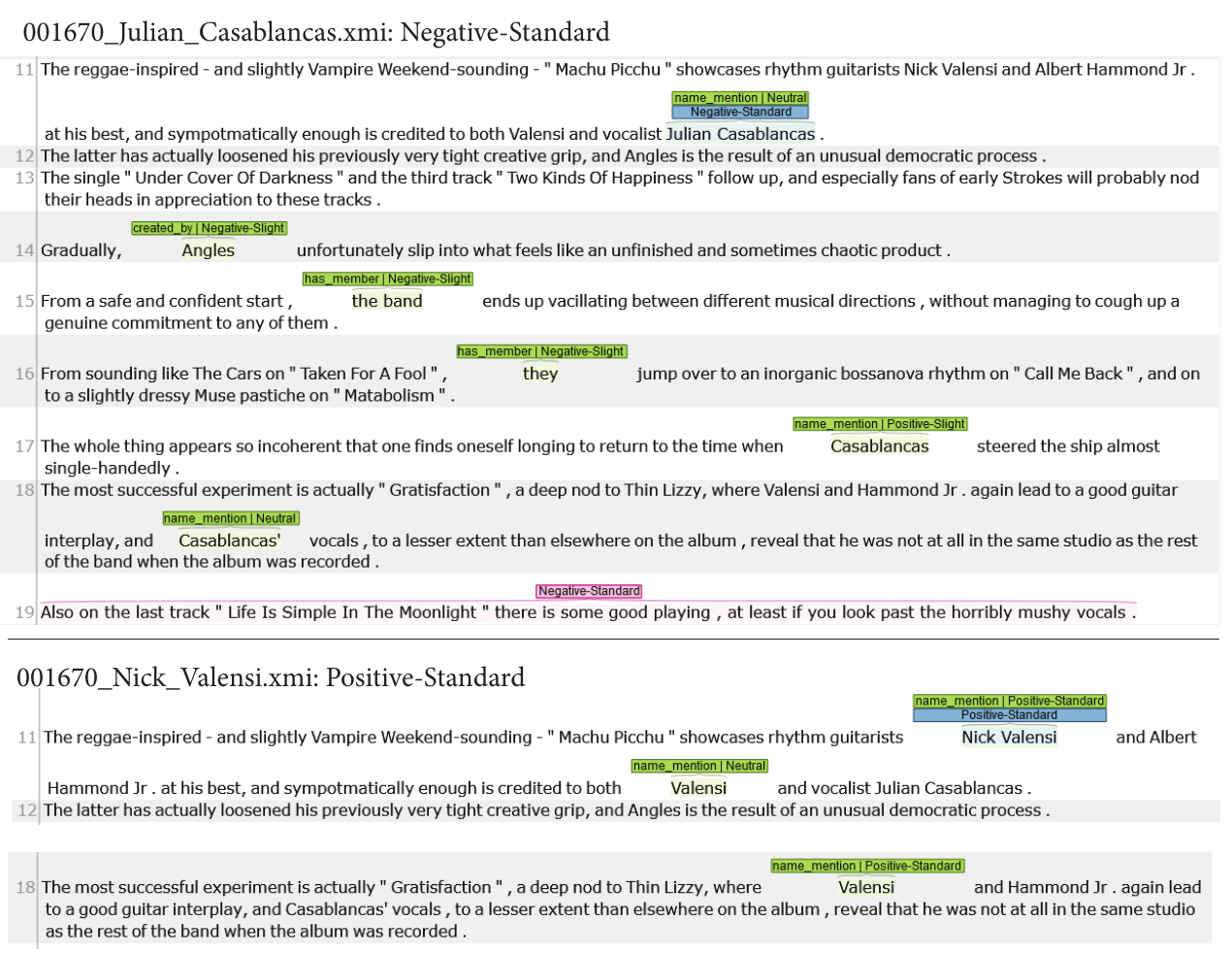}
	\caption{Annotations for two of the entities identified in the same text. Blue labels are document-level, green labels are segment-level, red labels are sentence-level. Sentence 19 is labeled as conveying a negative sentiment regarding Casablancas, since he is the vocalist.}
	\label{fig:inception}
\end{figure*}

\section{Baseline Details}
\label{sec:appendix_baseline}
Table \ref{tab:seq_label_params} shows the hyperparameters search space for the sequence labelling model we trained for predicting an entity's overall sentiment based on the sentiment expressed towards each entity mention. The code employed is a copy from the HuggingFace token classification task.

\begin{table} %
    \centering
    \small
    \begin{tabular}{ll}
    \toprule
    Parameter & Settings \\
    \midrule
     Models    &  NbAiLab/nb-bert base and large\\
     & \textbf{ltg/norbert3-large} \\
    Seeds     &  101, 202, 303\\
     Batch size    & \textbf{32}, 64\\
    Learning rate     & 1e-05, \textbf{5e-05}\\
    Epochs (best)    & 12 (\textbf{6})\\
    \bottomrule
    \end{tabular}
    \caption{Hyperparameters explored for fine-tuning a model for identifying and labeling name mentions and their polarity. Best options in bold. The narrow selection of models and hyperparameters is based on preliminary experiments with the material.}
    \label{tab:seq_label_params}
\end{table}

\section{Annotator Guidelines}\label{sec:appendix_guidelines}

These are the guidelines used for annotating the texts of \nrf for entity-level sentiment. This annotation was done as part of the ELSA project, using INCEpTION. The original guidelines are written in Norwegian. The following is a translation into English.

The guidelines are based on the guidelines for \nrf and the work that \nrf refers to:
Kauter, Marjan Van de et al. “The good, the bad and the implicit: a comprehensive approach to annotating explicit and implicit sentiment.” Language Resources and Evaluation 49 (2015): 685-720.

\subsection{Objective}
The main objective of the annotation is to create a dataset where sentiment expressed against entities in the document is annotated. "Entities" are limited to persons and organisations. First, the sentiment that the document as a whole expresses towards the entity is annotated, before the sentences that contribute to conveying sentiment towards the entity are derived. If possible, the recipient of the expressed sentiment in the sentence should be annotated with the sentiment and how this recipient relates to the main entity.
\paragraph{List of terms}

\begin{itemize}
    \item \textbf{Sentiment}: A positive or negative attitude towards something or someone.
    \item \textbf{Sentiment analysis}: An inference of the sentiment expressed in a text. This can occur both when the author conveys their sentiment directly, and when the author conveys statements or information that can be said to convey a positive or negative impression of the entity.
    \item \textbf{Annotate}: Labelling words or phrases and entering information about these items.
    \item \textbf{(Volitional) entity}: Individual people and groupings of people who have a proper name. This includes organisations, companies and parties. Geopolitical organisations, such as countries or cities, are also considered volitional entities where they function as actors with intent. Made-up characters and organisations are also volitional entities in the given text. Examples of volitional entities include "Elsa", "Beatles", "Jens Stoltenberg", "Black Widow", "Norske Skog" and "Oslo City Council". In this project, \textit{entity} is used as short form for volitional entity.
    \item \textbf{(Entity) mention}: Where an entity is mentioned either with all of or part of its proper name. In the text "Jens Stoltenberg came to visit. Stoltenberg seems tired at the moment." there are two mentions, "Jens Stoltenberg" and "Stoltenberg", where we can interpret it as both referring to the same entity.
    \item \textbf{Coreference}: Where an entity is mentioned without using the entity's proper name. This can be done by using nouns or pronouns such as "these", "the band", "the prime minister" or "he".
    \item \textbf{(Sentiment) target:} If a sentence expresses positivity or negativity towards something, the target is the word or words that represent this "something" that the sentiment is directed towards. 
    
\end{itemize}

\subsection{Degree of detail for the annotation}

The expressed sentiment should be directly related to the main entity we are annotating for. The annotation should take little consideration of domain knowledge, other than that which may be found in the text. Factual information should not be interpreted as carrying sentiment, unless a clear sentiment is also expressed. Irony and sarcasm where a negative sentiment can be expressed using otherwise positive words are annotated as negative sentiment.
The annotation distinguishes between 2 levels of intensity:

    \paragraph{"Standard"} is used where the sentiment is clear.
    \paragraph{"Slight"} is used where the sentiment is weaker in intensity. "Slight" is also used where the sentiment appears vague or uncertain.

\subsubsection{Document level}
For each entity, you must specify the sentiment that the document as a whole conveys towards this entity. This sentiment should be the annotator's impression of the document's sentiment towards the entity after reading, which is not necessarily an aggregate of the sentiment analysis at sentence and entity level.

The sentiment "Neutral" is used for all entities that are only mentioned in the text, without the text conveying any further sentiment towards the entity.
\subsubsection{Sentence level}

In cases where you find sentences that are relevant for conveying sentiment towards the entity, without finding a target that represents the entity or is related to the entity, the entire sentence should be annotated with the sentiment that is conveyed.
\subsubsection{Segment level}
At segment level, we annotate sentiment targets. The sentiment must appear in the same sentence as where the target is located. If it does not, the sentiment-bearing sentence should be annotated at sentence level.

For each annotated sentiment, the relationship to the sentiment target must be specified.

    \verb"-name_mention" is used where the entity is fully or partially mentioned by name. The name corresponds to the name of the main entity in the document.
    
    \verb"-anaphoric" is used where the entity has an anaphoric representation in the sentence through coreference.
    
    \verb"-is_member" is used where someone or something in the text is part of the main entity.
    
    \verb"-has_member" is used where the main entity is part of a larger group and the sentiment expressed towards the larger entity affects the sentiment towards the main entity.
    
    \verb"-created_by" is used where some kind of product is created by the main entity.

The word span that constitutes the sentiment target should be as short as possible, with the exception of proper names, where each part of the name that appears should be annotated together ("Barack Obama", not just "Barack" or just "Obama").

Where several possible sentiment targets appear in the relevant sentence, the following hierarchy is used to choose which relation to annotate:

\begin{enumerate} [ itemsep=1pt, topsep=2pt]
    \item  \verb"name_mention"
    \item \verb"anaphoric" 
    \item \verb"-is_member", \verb"has_member" or \verb"created_by"
    \item  Annotate at the sentence level
\end{enumerate}

If conflicting sentiment is expressed towards the same entity in the same sentence, the first representation of the entity (following the hierarchy above) should be annotated with the sentiment conveyed by the sentence as a whole.

\end{document}